# Computational Linguistics Meets Libyan Dialect: A Study on Dialect Identification

**Mansour Essgaer***
Department of Information Systems, Faculty of Information Technology, Sebha University, Sebha Libya
E-mail: man.essgaer@sebhau.edu.ly
ORCID iD: https://orcid.org/0000-0002-8447-5091
*Corresponding author

**Khamis Massud**
Department of Information Systems, Faculty of Information Technology, Sebha University, Sebha Libya
E-mail: km.ahmed@sebhau.edu.ly
ORCID iD: https://orcid.org/0009-0000-3267-9407

**Rabia Al Mamlook**
Business Administration, Trine University, Angola, Indiana, United States
E-mail: almamlookr@trine.edu
ORCID iD: https://orcid.org/0000-0002-2523-7819

**Najah Ghmaid**
Department of Computer Science, Faculty of Science, Sebha University, Sebha Libya
E-mail: naj.ghumeid@sebhau.edu.ly

**Abstract:** This study investigates logistic regression, linear support vector machine, multinomial Naive Bayes, and Bernoulli Naive Bayes for classifying Libyan dialect utterances gathered from Twitter. The dataset used is the QADI corpus, which consists of 540,000 sentences across 18 Arabic dialects. Preprocessing challenges include handling inconsistent orthographic variations and non-standard spellings typical of the Libyan dialect. The chi-square analysis revealed that certain features, such as email mentions and emotion indicators, were not significantly associated with dialect classification and were thus excluded from further analysis. Two main experiments were conducted: (1) evaluating the significance of meta-features extracted from the corpus using the chi-square test and (2) assessing classifier performance using different word and character n-gram representations. The classification experiments showed that Multinomial Naive Bayes (MNB) achieved the highest accuracy of 85.89% and an F1-score of 0.85741 when using a (1,2) word n-gram and (1,5) character n-gram representation. In contrast, Logistic Regression and Linear SVM exhibited slightly lower performance, with maximum accuracies of 84.41% and 84.73%, respectively. Additional evaluation metrics, including log loss, Cohen's kappa, and Matthew's correlation coefficient, further supported the effectiveness of MNB in this task. The results indicate that carefully selected n-gram representations and classification models play a crucial role in improving the accuracy of Libyan dialect identification. This study provides empirical benchmarks and insights for future research in Arabic dialect NLP applications.

**Index Terms:** Libyan Dialect, Dialect Classifications, Logistic Regression.

## 1. Introduction

The Arabic language is a complex, morphologically rich language with numerous dialects. More than thirty recognized dialects exist across the Arab world, including Modern Standard Arabic (MSA) and colloquial varieties such as Libyan, Egyptian, Levantine, Gulf, Iraqi, Yemeni, and Moroccan Arabic. While MSA serves as the primary form of formal written communication, dialects are predominantly used in everyday conversations. These dialects exhibit significant differences from MSA in phonology, morphology, lexical choice, and syntax, making their processing a challenging task for Natural Language Processing (NLP) applications [1].

With the increasing penetration of the internet in Arab countries, Arabic dialects have become more prevalent in digital communication, particularly on social media platforms such as Twitter, Facebook, and YouTube. Unlike structured text sources, social media content is characterized by high volume, velocity, and variety, presenting additional computational challenges. The informal and often noisy nature of social media text exacerbates the difficulty of processing dialectal Arabic, as it lacks standardized spelling and consistent linguistic structures [2]. To address these challenges, researchers have developed machine learning-based dialect identification techniques, which are crucial for various NLP applications, including machine translation, sentiment analysis, and content retrieval.

Libyan Arabic (LA), like many other Arabic dialects, lacks a standardized writing system and exhibits significant regional variation [3-4]. Despite the growing presence of Libyan dialectal content online, particularly with increased internet accessibility—from 14% in 2010 to 21.76% in 2019, according to World Bank indicators—there has been limited research on its computational processing [5]. Most existing studies focus on widely spoken dialects such as Egyptian and Saudi Arabic, leaving Libyan Arabic underrepresented in NLP research [6-8]. This research gap motivates the present study, which aims to develop an effective classification model for identifying Libyan Arabic dialects in social media text, specifically using Twitter as a data source. This task not only facilitates the creation of substantial linguistic resources but also is also critical for numerous NLP applications, including morphological analysis, sentiment analysis, topic detection, machine translation, and online content retrieval [7].

In this study, we evaluate various machine learning classification models for the identification of Libyan Arabic, trained on the Multi-Arabic Dialect Applications and Resources (QADI) corpus [4]. This dataset contains 540k sentences covering 18 Arabic dialects. Given the ongoing debate in the literature regarding the optimal approach for dialect classification, this study systematically assesses different models to determine the most effective technique for distinguishing Libyan Arabic from other dialects [9-11]. By addressing this gap, the study contributes to the development of robust Arabic dialect processing techniques and enhances NLP applications for Arabic social media content.

Despite this progress, Arabic dialects remain underexplored; particularly LD. Existing studies do not explicitly address the question: "What are the best NLP models for effectively distinguishing LD from other Arabic dialects?" To fill this gap. As a result, there is a need for systematic approaches to classify and analyse Libyan dialect utterances. This study addresses this gap by evaluating various machine-learning models and feature representations for Libyan dialect classification. Specifically, we investigate the effectiveness of different n-gram representations and meta-features to determine their impact on classification accuracy. The main contributions of this study are as follows:

- A comprehensive evaluation of meta-features extracted from a Libyan dialect corpus using the chi-square test, identifying relevant and non-relevant features.
- A comparative performance analysis of different classification models, including Logistic Regression, Linear SVM, Multinomial Naïve Bayes, and Bernoulli Naïve Bayes, using various n-gram feature representations.
- An empirical benchmark for Libyan dialect classification, providing insights into the most effective feature engineering and modeling strategies for dialect identification.

The findings of this study contribute to future advancements in NLP applications for the Libyan dialect, such as morphological analyzers, part-of-speech taggers, and machine translation. Effective feature selection and classification models can enhance dialect-specific linguistic analysis and translation between MSA and Libyan Arabic. Additionally, improved dialect identification supports applications like sentiment analysis and social media monitoring.

The remainder of this paper is organized as follows: section 2 is devoted to Libyan dialect linguistics facts, and section 3 is devoted to a literature review on language identification. We present the material and methods in section 4, and then in section 5, we present the result and discussion. and finally, section 6 is dedicated to the study conclusion.

## 2. Libyan Dialect Linguistic Fact

The Libyan Dialect (LD) is one of the Maghrebi dialects, spoken by approximately 7 million people. LD shaped by significant influences from Italian and to a lesser extent Turkish, as well as a profound impact from Berber. This linguistic mixture is evident across various levels, including lexicon, syntax, morphology, and phonology [12]. LD serves as a linguistic bridge between the dialects of the Arab East, characterized by their openness and simplicity, and the dialects of the Maghreb, which are uniform and phonetically closed.

Geographically, LD divided into three primary dialect regions. The eastern dialect, centered in Benghazi and Bayda, extends beyond Libya's borders into parts of Egypt. The western dialect, centered in Tripoli and Misrata, is distinct from the eastern variety. Additionally, there is a central and southern variety, centered on Sabha and El Jufra, which shares similarities with the western dialect. Despite these classifications, LD exhibits significant regional variation within Libya itself, with noticeable differences at various linguistic levels. Furthermore, LD differs substantially from Modern Standard Arabic (MSA), posing challenges for directly applying existing MSA-focused natural language processing (NLP) tools, which often yield suboptimal performance when used with LD.

Phonetically, LD retains several features associated with pre-Islamic Arab dialects. This includes distinct articulation point's letters. For instance, the Egyptian qaf (ق) is often pronounced as a jima or the Bedouin qaf, while in the middle and eastern regions, it may be realized as a gina. Additionally, the kaf (ك) can sometimes shift towards a sound close to shin (ش), and the ghain (غ) may merge with a pharyngealized variant of itself. Instances of phonetic assimilation also occur, where the pronunciation of weaker letters modified to align with stronger ones. For example, "Janzour" often pronounced as "Zanzour," "Jawz" (meaning 'two') becomes "Zouz," and "Al-Jazia" transforms into "Zaziya."

A unique characteristic of LD is its modest integration of foreign vocabulary, especially compared to other Maghrebi dialects. Italian influence is most notable in household terms (e.g., kishik = spoon, friketa = fork, cogina = kitchen, lavendino = washbasin), place names (e.g., aspetar = hospital, bag = paved road, garden = garden), and automotive terminology (e.g., freno = brake, chratori = accelerator, freno mano = hand brake, cochiniti = bearings, compressor = compressor). Despite these borrowings, LD is predominantly rooted in Arabic vocabulary.

Lexically, LD includes notable expressions and cultural markers. For instance, the word bahi (meaning "good" or "well") is widely recognized as a quintessential Libyan term. Additionally, certain antonymic terms used metaphorically, such as referring to coal as "whiteness," or calling the blind "the seer." In the eastern and central regions, "fire" is euphemistically referred to as "wellness." LD also features regionally specific words, such as ki (used in Gharyan to mean "stink" or "invalid"), which may be unfamiliar to other Libyans.

Phonological variation includes the substitution of certain sounds. For example, thaal (ذ), thaa (ث), and za (ظ) often replaced with dal (د), taa (ت), and daad (ض), respectively. Hence, words like dahab (gold), ta'lab (fox), and dahr (back) exemplify this transformation.

Another distinctive feature of LD is its extensive vocabulary. Nearly every noun, verb, or adverb has multiple synonyms, most of which originate from classical Arabic. Table 1 (below) illustrates key lexical differences across LD dialects for select terms.

Table 1. Words for libyan dialects from different regions of libya

| No | Word | Meaning | Tripolitania dialect | Eastern dialect | Central dialect |
|---|---|---|---|---|---|
| 1 | ملابس | Clothes | Hawaij حوايج | Dabish دبش | Malabis ملابس |
| 2 | اقفز | Jump | Aungoz انقز | Auzgob ازقب | Natt نط |
| 3 | ساخن | Hot | Nao نو | Hamoo حمو | Sakhana سخانة |
| 4 | انظر | look | Hagita حقيته | Shabhta شبحته | Shofta شوفته |
| 5 | اذهب | Go | Bara برا | Adi عدي | Eimshi امشي |
| 6 | كيس بلاستيك | Plastic bag | Porza بورزه | Shkara شكارة | Keis كيس |
| 7 | كثير | Too much | Halba هلبة | Wajed واجد | Wajed واجد |
| 8 | لماذا؟ | Why? | Alash علاش؟ | Leish ليش ؟ | Leish ليش؟ |
| 9 | رائحة | Smell | Sanah صنة | Banah بنة | Reiha ريحة |
| 10 | نحن | We | Eihni احني | Nahna نحنا | Nahna نحنا |
| 11 | مخدة | Pillow | Maghada مخدة | Wesada وسادة | Wesada وسادة |
| 12 | برد | cold | Sagah صقع | Sametheri سميطري | Sametheri سميطري |
| 13 | اجلس هنا | Sit here | Gamez eihni قمزاهني | Gamez hena قمزهنا | Gamez hena قمزهنا |
| 14 | الاكل | Food | Makela ماكلة | Wekal وكال | Khashera خشرة |
| 15 | كذاب | Liar | Kathab كذاب | Kathab كذاب | Balaot بلعوط |

## 3. Related Works

There have been numerous attempts to create Arabic dialects applications. the latest trends involve the use of machine learning approaches, particularly neural networks, and the development of corpus that are specific to Arabic dialects. These trends aim to address the challenges posed by the high level of dialectal variation in Arabic, which increases the ambiguity in identifying the origin of a spoken utterance or written text.

There have been numerous attempts to create Arabic dialects applications. They employ various approaches and techniques. Here, we review and present the most relevant to our work. In terms of identifying Arabic dialects, some studies have come up with mixed finding: [13] found that the pre-trained system based outperform other models in identifying Arabic dialects, and [14] found that prosodic features can be used to automatically identify Arabic dialects. However, [15] found that no clear border can be found between the dialects, but a gradual transition between them, suggesting that it is difficult to automatically identify Arabic dialects. [16] found that no clear border can be found between the dialects, but a gradual transition appears between them. This is directly relevant to our research question. It suggests that it is difficult to automatically identify Arabic dialects.

Many researchers addressed the gap of the lack of a dialectical Arabic corpus for NLP. However, in the last ten years, progress in this regard start to gain much attention. As a result, there is an increasing interest in NLP tools for written Arabic dialects, such as sentiment analysis to decide what customers think about goods. [6] presented a manually annotated large corpus, created from Twitter, significant efforts were conducted to crawl Twitter for dialectical Arabic. [8] propose a social sentiment analysis for the Egyptian dialect dataset gathered from Twitter, containing ten thousand tweets, which are manually classified as objective, subjective positive, subjective negative, and subjective mixed. [17] describe a method to collect dialectal speech from YouTube videos to create a large-scale Dialect Identification dataset.

[3] present QADI, an automatically collected dataset of tweets belonging to a wide range of country-level Arabic dialects covering 18 different Arab countries. [18] present the results of the First Nuanced Arabic Dialect Identification Shared Task (NADI). [19] present the findings and results of the Second Nuanced Arabic Dialect Identification Shared Task.

Although Arabic dialects have gained much more attention, there have been a few works on the LD. [9] have introduced the first LD corpus, which includes 5K statements collected from Twitter. The corpus is annotated manually into fifteen categories dominated by sports, which could be taken into account when using this corpus. [20] work to focus on building a lexicon-based sentiment analysis system, which is composed of a set of weighted adjectives and adverbs, collected manually from the Libyan tweet corpus. [21] suggests that machine learning could be used to study code-switching between Tamazight and Arabic in Libyan Berber news broadcasts. [22] is an annotated bibliography of Arabic and Berber in Libya, which could be used as a resource for studying LD.

Recent studies have focused on developing resources and models for processing LD. [9] constructed a Twitter-based LD corpus, facilitating dialect identification and sentiment analysis. [23] developed a polar-oriented LD corpus using an emoji-based lexicon, highlighting the role of emojis in sentiment analysis. [24] employed a hybrid Convolutional Neural Network and Meta-learning approach to classify cyberbullying in Arabic comments, achieving high accuracy rates. [25] utilized Named Entity Recognition to identify entities related to illegal migration in Libya, aiding policymakers and researchers, the corpus collected mix with LD. [26] applied machine learning models to predict customer satisfaction for Libyan telecom companies based on Twitter data, providing insights for service improvement. These studies collectively advance NLP applications for LD, addressing challenges like code-switching, orthographic variations, and resource scarcity.

Processing LD for NLP presents challenges such as code-switching between Arabic dialects and foreign languages, orthographic variations due to the absence of standardized writing systems, and limited standardized resources. Addressing these issues is crucial for developing effective NLP tools tailored to LD.

## 4. Material and Methods

The proposed system includes five stages: data collection and corpus characteristics, data preprocessing and engineering, feature extraction, feature selection, classification and evaluation. Fig. 1 depicts an overview of the proposed system.

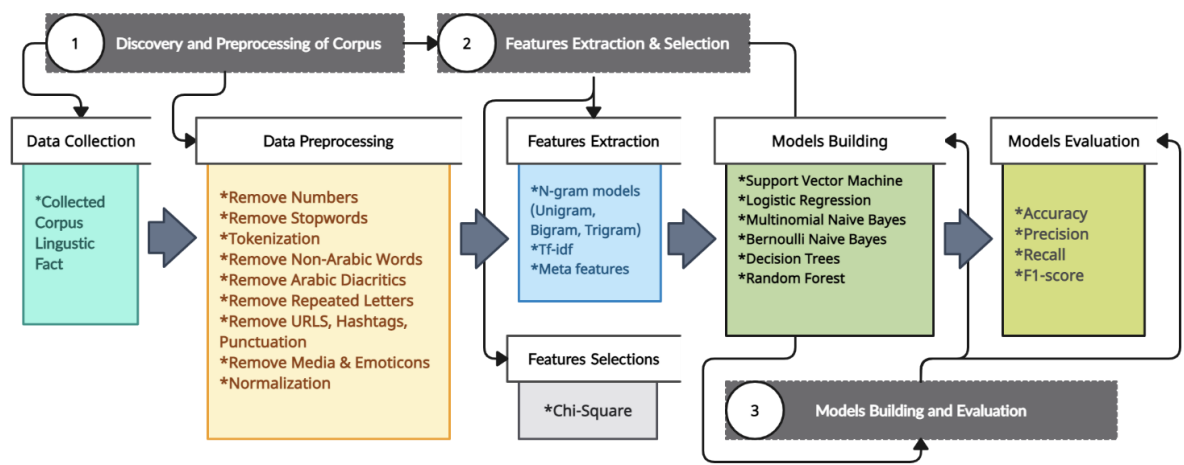

Fig.1. Overview of the proposed system

### 4.1. Collected Corpus Linguistic Facts

In our research, we utilized the QCRI Arabic Dialects Identification (QADI), as introduced in the study by [4]. QADI consists of a collection of 18 different Arab dialects. This corpus encompasses 458,197 sentences, distributed among the 18 dialects. Figure 2 (left) illustrates the sentence distribution of each dialect by country. Notably, the Libyan dialect ranks 4th, comprising approximately 8% of the total dialect distributions, amounting to around 36,000 sentences. To gain further insights into the positioning of the Libyan dialect relative to other dialects, we calculated the distinct vocabulary for each dialect. The Libyan dialect ranks 4th in terms of unique vocabulary as shown in Figure 2 (Right), with approximately 47,000 distinct words. The analysis of distinct vocabulary proves valuable in distinguishing between the different dialects.

According to [27] Tunisian and Libyan dialects demonstrate similarities owing to their geographical proximity and historical interactions. Based on this understanding, when selecting a suitable sample for training a model, it would be preferable to bias the selection towards including more Tunisian dialect. The rationale behind this bias is the assumption that the model would likely perform better when trained on a dataset that encompasses dialects with greater similarities. As a result, we have chosen to include approximately 5000 Tunisian dialect sentences, while the remaining sentences randomly selected from other Arabic dialects.

Figure 3 provides an overview of the most frequent unigram, bigram, and trigram tokens in the Libyan dialect. Analysis of the figure reveals that the most common word considered a non-discriminatory stop word because it commonly employed in Libyan and other Arabic dialects. Consequently, a decision has been made to remove specific tokens, such as (الله، والله، ربي) (referring to the name of God), (الله يبارك فيك) as a trigram token, and its variant in the bigram token (يبارك فيك، الله يبارك). Similar measures are recommended for other tokens, including (حسبي الله ونعم الوكيل) and its variation (حسبي الله، ونعم الوكيل), and so forth. In Figure 4, the sentence length of the Libyan and the other dialects combined depicted. The analysis reveals that the average sentence length in both Libyan and other dialects is nearly identical. However, the Libyan dialect exhibits a notable number of sentences with a density ranging between 4000 and 5000 with a length of 50 compared to the other dialects.

The Entropy Shift diagram shown in figure 5 analyzes changes in entropy across different data groups to identify the factors contributing most to overall entropy. Entropy measures uncertainty in data, with higher values indicating greater diversity. The graph shows how entropy changes when shifting between groups, highlighting the most influential factors. Comparing "Libyan" and "Other" groups, the average entropy shift is 14.17 and 13.74, respectively. The slightly higher shift for "Libyan" suggests it contributes more to the variability of the "center" variable.

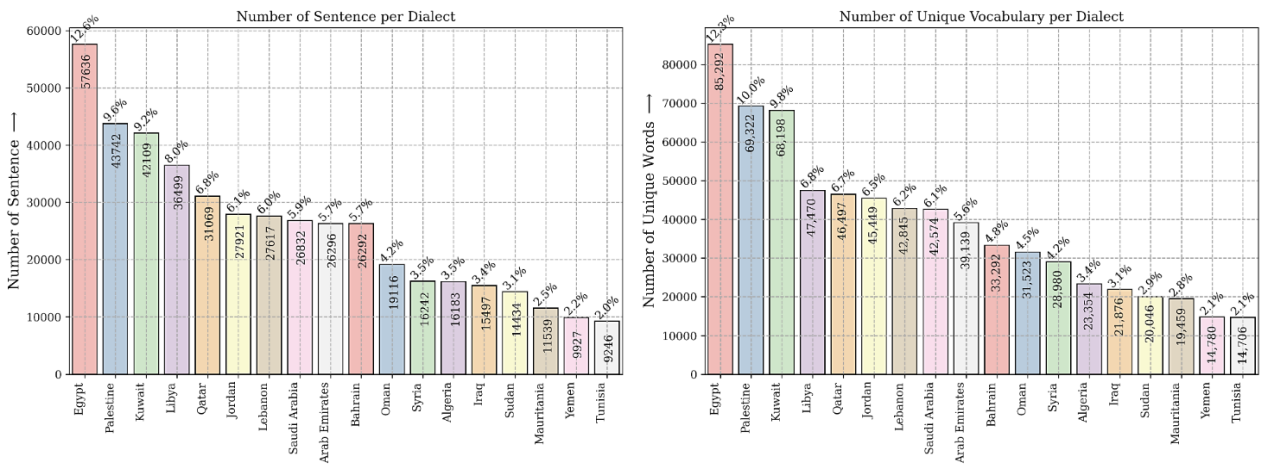

Fig.2. The number of sentences of each dialed (left), the unique vocabulary of each dialect (right)

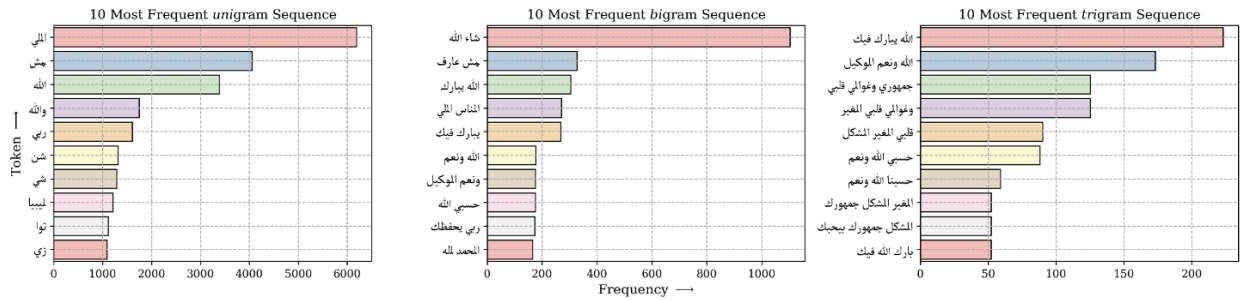

Fig.3. The ten most frequent (uni, bi, tri) gram tokens used in the Libyan dialect

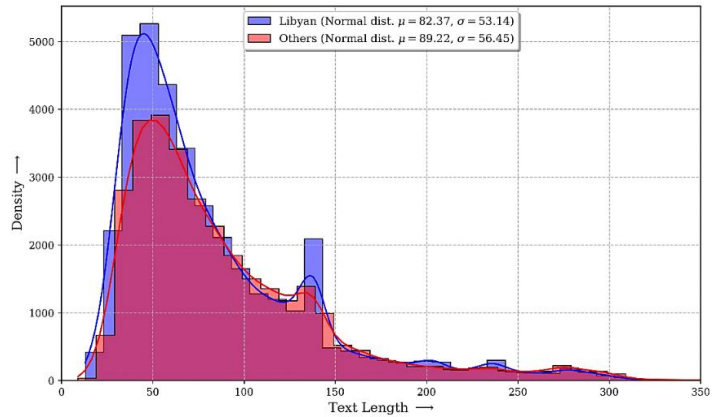

Fig.4. The length of Libyan sentences compared to other dialects

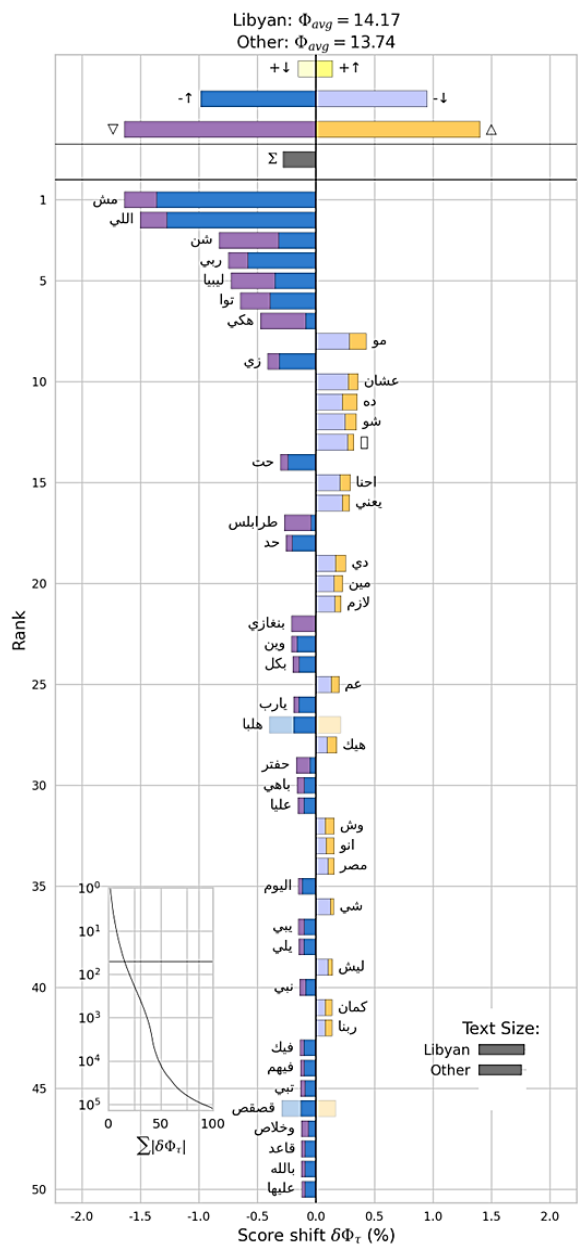

Fig.5. Entropy shift for the Libyan dialect compared to other

## 4.2. Features Indicative of Informal Sentence Characteristics (Meta Features)

In addition to analyzing the dialectalness of words, this article explores additional features that can provide insights into the informal nature of a sentence listed in Table 2. By considering these features, researchers and language professionals can gain a more comprehensive understanding of sentence informality, leading to improved text analysis and comprehension.

- Percentage of word length, character length, and space count: The first set of features that can help estimate the informality level of a sentence includes the percentage of word length, character length, and space count. Longer words and sentences with a higher character count and fewer spaces may suggest a more informal writing style.
- Punctuation usage: Punctuation marks can play a significant role in determining sentence informality. Features such as the presence of exclamation marks, happy or sad emoticons, and the usage of periods and commas can indicate a more casual or informal tone.
- Presence of contact information and social media indicators: Another set of features to consider involves the presence of contact information and social media indicators within a sentence. These can include phone numbers, emails, mentions, hashtags, and other markers commonly associated with informal communication channels.

Table 2. Features indicative of informal sentence characteristics

| Feature | Description |
|---------|-------------|
| Percentage Word Length | The percentage of word length relative to the total sentence length. |
| Character Length | The total number of characters in the sentence, excluding spaces. |
| Space Count | The number of spaces present in the sentence. |
| Punctuation Usage | Binary features indicate the presence (yes) or absence (no) of punctuation marks such as periods, commas, exclamation marks, and emoticons. |
| Contact Information (Phone number, Email) | Binary feature indicating the presence (yes) or absence (no) of phone numbers and email addresses. |
| Social Media Indicators (Mentions, Hashtags) | Binary features indicate the presence (yes) or absence (no) of social media indicators such as mentions and hashtags. |

### 4.3. Feature Extraction

This study utilizes an N-gram model, where tokens are represented as fixed-length word sequences. The model includes unigrams, bigrams, and trigrams. Table 3 presents the number of tokens for each N-gram type.

Table 3. Number of tokens for each n-gram model

| N-grams model | Tokens | Words | Hapax legomenon |
|---------------|--------|-------|-----------------|
| Unigram | 199003 | 1022966 | 132377 |
| Bigram | 758759 | 949971 | 695415 |
| Trigram | 851664 | 876994 | 838183 |

The table provides insights into the number of unique tokens, the overall number of tokens, and the count of hapax legomena, which refers to words that appear only once in the dataset. These statistics can be useful for understanding the vocabulary richness, token distribution, and the prevalence of rare or unique word sequences within the dataset analyzed in the study.

To assess word importance, the term frequency-inverse document frequency (tf-idf) measure is employed. It assigns weights based on word frequency within a document, with higher frequencies indicating stronger associations with specific categories. Term frequency represents word occurrences within a document, while inverse document frequency captures overall word significance. Calculating tf-idf scores generates feature importance for classification.

However, the tf-idf vectorized representation loses syntactical information. To address this drawback, the study considers n-grams at both word and character levels. Consequently, unigram and bigram word-level tf-idf vectors are extracted, along with character-level tf-idf vectors using n-gram values ranging from 1 to 5. This approach mitigates the loss of word order details, resulting in a more comprehensive representation that enhances the analysis of feature importance and improves the classification process.

### 4.4. Preprocessing Steps

To improve the quality of textual data and enhance model performance, several preprocessing steps were applied. These steps addressed common challenges in Arabic dialect processing, particularly for informal social media texts.

- The first step involved removing media elements, emoticons, and punctuation. Social media texts frequently contain emojis, images, and excessive punctuation, which do not contribute to dialect identification. For instance, a sentence like "الجو حموو اليوم 😊 !!!" was cleaned to "الجو حمو اليوم", ensuring that the text remained meaningful without unnecessary distractions.
- Next, URLs, numbers, and hashtags were removed. Social media users often include links and hashtags that do not carry linguistic significance for classification. For example, a tweet such as "شوفتو المباراة اليوم! #ليبيا فازت 🏆 تابع هنا: www.example.com" was processed into "شوفتو المباراة اليوم ليبيا فازت", removing both the URL and the hashtag while retaining the core message.
- Normalization was then applied to standardize different forms of Arabic letters and remove diacritics. Arabic characters can have multiple representations, which can introduce inconsistencies in analysis. The normalization process replaced various forms of the letter alef (e.g., أ and إ) with a simple ا, converted taa marbutah (ة) into haa (ه), and transformed alef layyinah (ى) into yaa (ي). For example, the phrase "إلّي صار مايتعاود!" was normalized to "اللي صار مايتعاود", ensuring consistency in character usage.
- Non-Arabic characters, including Latin letters, numbers, and special symbols, were also removed. This was particularly important since Arabic social media users sometimes mix languages. For instance, the phrase "Hello! شن حالكم؟ 😊" was cleaned to "شن حالكم", eliminating the English word while preserving the relevant Arabic content.
- Another critical step was handling letter repetition, a common feature in Libyan dialect writing, where users extend letters for emphasis. To standardize this, repetitions beyond two consecutive letters were reduced. For example, "حااار هلباااا!!!!!!!!" was processed into "حار هلبا", maintaining readability while preventing exaggerated spellings from affecting classification accuracy.

- Tokenization was then performed to split sentences into individual words or tokens, which facilitates feature extraction for machine learning models. A sentence like "شن حالك اليوم؟" was tokenized into ["اليوم" ,"حالك" ,"شن"], allowing each word to be analyzed separately.
- Finally, stop words were removed to reduce noise and focus on meaningful words. Arabic stop words include common pronouns, conjunctions, and prepositions that do not add significant value to classification tasks. For example, the sentence "أنا في طرابلس اليوم ومشيت للسوق" was processed into "طرابلس مشيت سوق", where words like أنا (I) and في (in) were removed. This step improved classification by emphasizing content words rather than frequently occurring function words.

By applying this comprehensive preprocessing pipeline, the dataset was refined to retain dialect-specific linguistic features while minimizing irrelevant elements. These steps enhanced the model's ability to distinguish the Libyan dialect more effectively, contributing to improved classification performance.

*4.5. Classifiers*

We present a comparative analysis of six classifiers: Linear Support Vector Machine, Logistic Regression, Multinomial Naive Bayes, and Bernoulli Naive Bayes, Random Forest, and Decision Tree. The objective of this research is to evaluate the performance of these classifiers and understand their suitability for on the Libyan dialect.

- Linear Support Vector Machine (SVM): is a powerful and widely-used supervised learning algorithm. It aims to find the best hyperplane that separates data points into different classes. It is known for its ability to handle high-dimensional data and its robustness against overfitting.
- Logistic Regression: is a statistical modeling technique used for binary classification problems. It estimates the probability of a certain event occurring based on input variables. Logistic regression is particularly useful when dealing with categorical outcomes and provides interpretable results.
- Multinomial Naive Bayes: is a probabilistic classifier commonly used for text classification. It assumes that features are conditionally independent and follows a multinomial distribution. This classifier is well-suited for problems involving discrete features, such as word frequencies in text analysis.
- Bernoulli Naive Bayes: is another variant of Naive Bayes that assumes binary features. It is often used for document classification tasks, where the presence or absence of certain words is considered as features. Bernoulli Naive Bayes works well with sparse datasets.
- Decision Trees: are simple yet effective binary classifiers that use a tree-like structure to make decisions based on feature values. They are interpretable and can handle both categorical and numerical features.
- Random Forest: is an ensemble learning method that combines multiple decision trees to make predictions. It can be used for binary classification in text analysis by aggregating the predictions of individual trees.

*4.6. Performance Evaluation*

To evaluate the performance of these classifiers, various metrics such as accuracy, precision, recall, F1-score, log loss, Cohen's kappa score, and Matthew's correlation coefficient will be utilized.

## 5. Result and Discussion

We have built several systems to achieve the models described above, whereby two experiments in conducted: 1, to evaluate the meta-features extracted from the corpus, 2, classifier performance evaluation using different n-gram model using words and characters.

*5.1. Meta Features and their Significance*

Multiple meta-features have been generated in this comprehensive study. These features do not directly pertain to the dialectal Ness of words within a sentence; instead, they serve to estimate the level of informality exhibited. To identify the most effective meta-features, we conducted rigorous experiments employing the chi-square approach. This statistical test allows us to measure the association between the occurrence of each meta-feature and the informality of sentences. A critical level of 0.05 was chosen for our analysis. The results of each feature evaluation are illustrated in Figure 6.

Based on the chi-square analysis, it was observed that the emails, mention, sad emotion, and happy emotion counts did not exhibit significant association with the informality of sentences. These features did not contribute significantly to the identification of dialectalness, failing to surpass the critical level of 0.05. Therefore, we have decided to exclude these features from our subsequent investigations and focus on the more influential meta features.

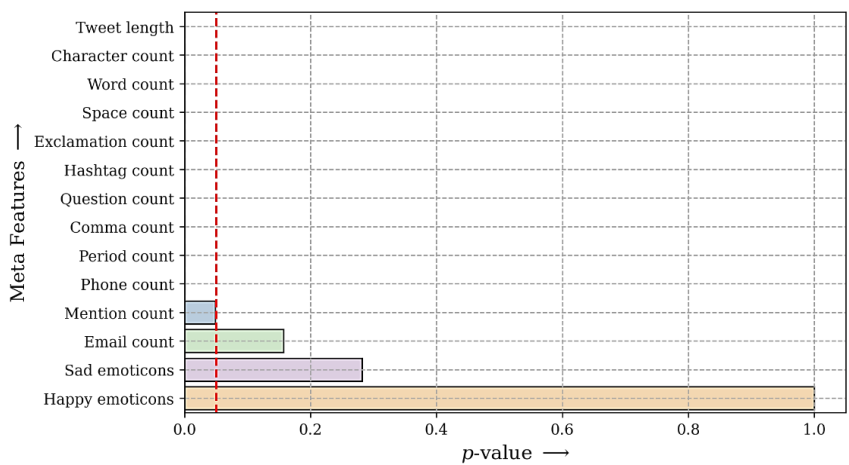

Fig.6. Chi-Square test results of the Meta features based on p-value

## 5.2. The Performance of the Different Classification Models

The results of different classification methods for Arabic dialect identification are presented in the table 4. Logistic Regression, Linear SVM, Multinomial NB, and Bernoulli NB classifiers were evaluated using various word and character n-gram feature combinations.

Among the Logistic Regression models, the (1, 1) word n-gram with (1, 5) character n-gram achieved the highest accuracy of 84.41% with an F1-score of 0.84458. The Linear SVM models also performed well, with the (1, 2) word n-gram and (1, 5) character n-gram combination achieving an accuracy of 84.73% and an F1-score of 0.84580. Multinomial NB models showed consistent performance across different n-gram combinations, with accuracies ranging from 85.21% to 85.89% and F1-scores ranging from 0.85614 to 0.85741. The (1, 2) word n-gram and (1, 5) character n-gram combination achieved the highest accuracy and F1-score among the Multinomial NB models. In contrast, the Bernoulli NB models had lower accuracies compared to other classifiers, with accuracies ranging from 83.99% to 84.22% and F1-scores ranging from 0.83173 to 0.83415. The (1, 2) word n-gram and (1, 5) character n-gram combination yielded the highest accuracy and F1-score among the Bernoulli NB models.

Overall, the results indicate that different classification methods perform differently for Arabic dialect identification. Logistic Regression, Linear SVM, Multinomial NB, and certain Bernoulli NB models showed promising results in terms of both accuracy and F1-score. Further analysis and experimentation can be conducted to fine-tune the models and explore other feature combinations for improved accuracy.

Table 4. Performance of the proposed methos based on accuracy, precision, recall and F1 score

|  | Word gram | Character gram | Accuracy | Precision | Recall | F1 |
|---|---|---|---|---|---|---|
| Logistic Regression | (1, 1) | (1, 5) | 0.84205 | 0.84712 | 0.84712 | 0.84458 |
| Logistic Regression | (1, 2) | (1, 5) | 0.84137 | 0.84534 | 0.83562 | 0.84045 |
| Logistic Regression | (1, 3) | (1, 5) | 0.83890 | 0.84380 | 0.83178 | 0.83775 |
| Linear SVM | (1, 1) | (1, 5) | 0.84342 | 0.84352 | 0.84329 | 0.84340 |
| Linear SVM | (1, 2) | (1, 5) | 0.84726 | 0.85395 | 0.83781 | 0.84580 |
| Linear SVM | (1, 3) | (1, 5) | 0.84740 | 0.85618 | 0.83507 | 0.84549 |
| Multinomial NB | (1, 1) | (1, 5) | 0.85521 | 0.84657 | 0.86767 | 0.85699 |
| Multinomial NB | (1, 2) | (1, 5) | 0.85589 | 0.84844 | 0.86658 | 0.85741 |
| Multinomial NB | (1, 3) | (1, 5) | 0.85425 | 0.84517 | 0.86740 | 0.85614 |
| Bernoulli NB | (1, 1) | (1, 5) | 0.83986 | 0.87625 | 0.79151 | 0.83173 |
| Bernoulli NB | (1, 2) | (1, 5) | 0.84151 | 0.87853 | 0.79260 | 0.83336 |
| Bernoulli NB | (1, 3) | (1, 5) | 0.84219 | 0.87894 | 0.79370 | 0.83415 |

The error analysis from the confusion matrix in Figure 7 highlights key misclassification trends among the models. Multinomial NB, despite achieving the highest accuracy (85.89%), shows a slightly higher rate of false positives compared to Linear SVM and Logistic Regression, suggesting that it may misclassify non-Libyan dialect samples as Libyan more often. Bernoulli NB, while having the highest precision (87.8%), exhibits the lowest recall (79.3%), indicating that it struggles to correctly identify Libyan dialect samples, leading to more false negatives. Logistic Regression and Linear SVM show balanced performance, with Linear SVM (1,2 word n-gram, 1,5 character n-gram) achieving the best trade-off between precision and recall, minimizing both false positives and false negatives. The impact of n-gram features is also evident, as variations in word and character n-gram combinations influence model performance.

These findings emphasize the trade-off between capturing dialect variations and avoiding misclassifications, which is crucial for improving dialect identification accuracy.

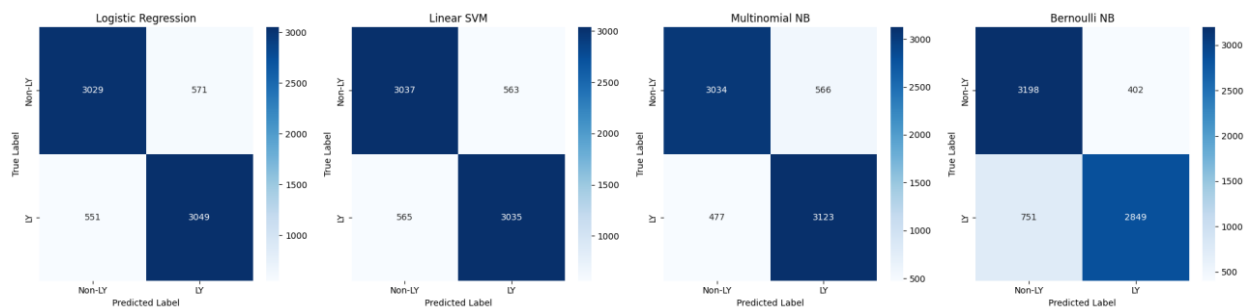

Fig.7. Confusion matrix for the different classification methods

Given the marginal differences in the previous table, we have included another metric shown in Table 5 for further analysis. The table presents the results of different classifiers using three-evaluation metrics, namely log loss, Cohen's kappa score, and Matthew's correlation coefficient, are provided to assess the performance of each classifier.

Looking at the log loss values, it can be observed that Multinomial NB with (1, 2) word grams and (1, 5) character grams achieved the lowest log loss of 4.97743, indicating better probability estimation and calibration compared to other classifiers. Moving on to Cohen's kappa score, which measures the agreement between predicted and true class labels, Multinomial NB consistently performs well across different gram models. It obtains the highest Cohen's kappa score of 0.71178 with (1, 2) word grams and (1, 5) character grams, indicating a strong level of agreement beyond what is expected by chance. Similarly, when considering the Matthews correlation coefficient, which takes into account true positives, true negatives, false positives, and false negatives, Multinomial NB exhibits the highest values across different gram models. It achieves a Matthews correlation coefficient of 0.71194 with (1, 2) word grams and (1, 5) character grams, indicating a well-balanced measure of classification performance. Overall, Multinomial NB consistently performs well across all three evaluation metrics, suggesting its effectiveness in the dialect identification task.

Table 5. Performance of the proposed methods based on log loss Cohen kappa score and Matthews corr. coeff score

|  | Word gram | Character gram | log loss | Cohen kappa score | Matthews Corr. Coeff. |
|---|---|---|---|---|---|
| Logistic Regression | (1, 1) | (1, 5) | 5.38433 | 0.68822 | 0.68823 |
| Logistic Regression | (1, 2) | (1, 5) | 5.47895 | 0.68274 | 0.68278 |
| Logistic Regression | (1, 3) | (1, 5) | 5.56412 | 0.67781 | 0.67788 |
| Linear SVM | (1, 1) | (1, 5) | 5.40798 | 0.68685 | 0.68685 |
| Linear SVM | (1, 2) | (1, 5) | 5.27550 | 0.69452 | 0.69464 |
| Linear SVM | (1, 3) | (1, 5) | 5.27077 | 0.69479 | 0.69501 |
| Multinomial NB | (1, 1) | (1, 5) | 5.00109 | 0.71041 | 0.71063 |
| Multinomial NB | (1, 2) | (1, 5) | 4.97743 | 0.71178 | 0.71194 |
| Multinomial NB | (1, 3) | (1, 5) | 5.03421 | 0.70849 | 0.70874 |
| Bernoulli NB | (1, 1) | (1, 5) | 5.53098 | 0.67973 | 0.68293 |
| Bernoulli NB | (1, 2) | (1, 5) | 5.47420 | 0.68301 | 0.68630 |
| Bernoulli NB | (1, 3) | (1, 5) | 5.45055 | 0.68438 | 0.68763 |

The Receiver Operating Characteristic (ROC) curve is a graphical representation that depicts the performance of a binary classifier by plotting the true positive rate against the false positive rate at different classification thresholds. In Figure 8, the ROC curve for the utilized classifier is presented, revealing a marginal difference. This indicates that the classifiers, such as Logistic Regression and Linear SVM, exhibit comparable performance in terms of true positive and false positive rates, as evidenced by an Area Under the Curve (AUC) value of 0.92. Similarly, Multinomial NB and Bernoulli NB demonstrate similar performance with an AUC of 0.93. Overall, these results suggest that the classifiers perform similarly in distinguishing between positive and negative classes.

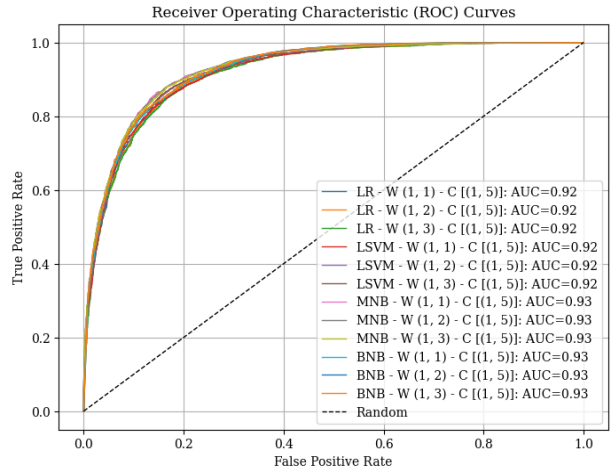

Fig.8. Receiver operating characteristics for the different classification methods

## 6. Conclusions

This study provides valuable insights into the performance of various classifiers and feature representations for dialect identification. The evaluation of different classification methods using word grams and character grams reveals the strengths and limitations of each approach. The metrics, including accuracy, precision, recall, F1-score, log loss, Cohen's kappa score, and Matthew's correlation coefficient, shed light on the effectiveness of the models in distinguishing between Libyan and other Arabic dialects. When considering the classification performance, Logistic Regression, Linear SVM, Multinomial NB, and Bernoulli NB consistently demonstrate competitive results across different n-gram models. These models exhibit similar performance in terms of accuracy and F1-score, indicating their effectiveness in correctly classifying dialects. However, it is important to note that there are slight variations in their performance metrics, such as log loss, Cohen's kappa score, and Matthew's correlation coefficient, which provide additional insights into the models' predictive capabilities. The use of n-gram models, including unigrams, bigrams, and trigrams, allows for a comprehensive analysis of the token distribution, vocabulary richness, and word sequence prevalence in the dataset. The consideration of both word and character-level n-grams helps alleviate the limitation of losing word order information in the tf-idf vectorized representation. This enhances the feature importance and contributes to better classification performance. Overall, this study highlights the importance of selecting appropriate classifiers and feature representations for dialect identification tasks. The results serve as a foundation for further research in exploring alternative approaches, feature engineering techniques, and modeling strategies to enhance the accuracy and robustness of Arabic dialect identification systems

**Authors' Profiles**

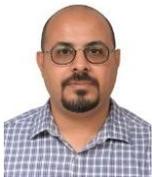

**Mansour Essgaer**, Ph.D. in Computer Science from the National University of Malaysia (Universiti Kebangsaan Malaysia) in 2016, specializing in artificial intelligence. He served as a Researcher at the National University of Malaysia's Center for Artificial Intelligence Technology from October 2011 to January 2016. Since 2017, he has been an Assistant Professor in the Faculty of Information Technology at Sebha University, Libya. His research interests include data mining, artificial intelligence, machine learning, natural language processing, and combinatorial optimization. He has published several articles in international journals and conferences, contributing to advancements in his field. Essgaer is a member of the Institute of Electrical and Electronics Engineers (IEEE) and actively participates in research and professional activities.

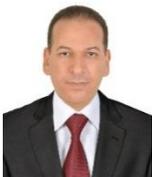

**Khamiss M. S. Ahmed**, Ph.D. in Computer Science (Artificial Intelligence, Pattern Recognition), from Faculty of information science and technology, National University of Malaysia. University professor and researcher specializing in artificial intelligence, in the field of Pattern Recognition, I work as an associate professor, Faculty of Information Technology, University of Sebha, Libya. I work as a director of the Graduate Studies Office at the college; I supervise many theses for graduate students. I was also selected as a confidential evaluator as well as an internal and external examiner for many scientific theses. I published many research papers in international journals and conferences. I organized many workshops and public lectures related to the field of artificial intelligence as well as scientific research methods and methodologies. And exploiting the capabilities of generative artificial intelligence in various scientific research activities.

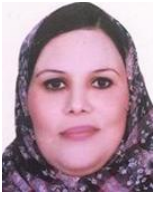

**Rabia Al Mamlook,** is a data scientist with a robust background in industrial engineering and engineering management. He earned his Ph.D. in Industrial Engineering and Engineering Management from Western Michigan University (WMU), USA. He also holds a Master's in Applied Statistics & Biostatistics from WMU and a Master's in Engineering Management from the University of Tripoli, Libya. His expertise encompasses machine learning, data mining, statistical process quality control, and data visualization, applied to large datasets in industries and healthcare engineering. Proficient in programming languages such as R, SAS, Minitab, and Python, his research interests include smart manufacturing, data models, and deep learning.

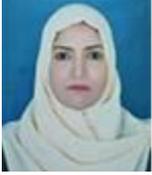

**Najah Ghumeid** was born in Sebha, southern Libya, in 1985. She received the B.S. degree in computer science from the faculty of sciences, Sebha University, Libya, in 2008. She is currently a lecturer and M.S. candidate in artificial intelligence and machine learning at the College of Information Technology, Sebha University, with an expected degree completion in 2025. She co-authored the paper "Addressing the Libyan Arabic Dialect Identification: A Comparative Study of Ensemble Classification Methods" (Tripoli, Libya, IEEE-MI-STA 4th, 2024). Professionally, she serves as an accredited trainer at Sebha University's Training and Development Centre, specializing in programs such as IC³, ICDL, MOS, and TOT. She has over 10 years of experience designing and accrediting IT training curricula, including foundational contributions to the Centre's inaugural programs in 2016. Previously, she led the implementation of the university's staff classification system (2016) and served on its Financial Entitlements Committee (2019). Her research focuses on natural language processing (NLP) and artificial intelligence (AI) systems. Ms. Ghumeid holds certifications as a Microsoft Office Specialist (MOS, Certiport, 2013), IC³ Global Standard 4 Instructor (Certiport, 2010), ICDL Master Instructor (2013), and Trainer of Trainers (TOT, American Canadian Board, 2016).